\documentclass[sigconf]{aamas}

\usepackage{balance} 

\usepackage{algorithm}      
\usepackage{algorithmic}    
\usepackage{enumerate}
\usepackage{enumitem}

\doi{BELB5985}

\makeatletter
\gdef\@copyrightpermission{
  \begin{minipage}{0.2\columnwidth}
   \href{https://creativecommons.org/licenses/by/4.0/}{\includegraphics[width=0.90\textwidth]{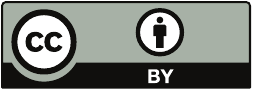}}
  \end{minipage}\hfill
  \begin{minipage}{0.8\columnwidth}
   \href{https://creativecommons.org/licenses/by/4.0/}{This work is licensed under a Creative Commons Attribution International 4.0 License.}
  \end{minipage}
  \vspace{5pt}
}
\makeatother

\setcopyright{ifaamas}
\acmConference[AAMAS '26]{Proc.\@ of the 25th International Conference
on Autonomous Agents and Multiagent Systems (AAMAS 2026)}{May 25 -- 29, 2026}
{Paphos, Cyprus}{C.~Amato, L.~Dennis, V.~Mascardi, J.~Thangarajah (eds.)}
\copyrightyear{2026}
\acmYear{2026}
\acmDOI{}
\acmPrice{}
\acmISBN{}

\title[AAMAS-2026 Formatting Instructions]{IPD: Boosting Sequential Policy with Imaginary Planning Distillation in Offline Reinforcement Learning}

\author{$\text{Yihao Qin}^{*}$}
\thanks{\textsuperscript{*}\text{Equal contribution;} \textsuperscript{\dag}\text{Corresponding author}}
\affiliation{
  \institution{The Hong Kong University of
Science and Technology (Guangzhou)}
  \city{Guangzhou}
  \country{China}}
\email{yqin637@connect.hkust-gz.edu.cn}

\author{$\text{Yuanfei Wang}^{*}$}
\affiliation{
  \institution{Peking University}
  \city{Beijing}
  \country{China}}
\email{yuanfei_wang@pku.edu.cn}

\author{Hang Zhou}
\affiliation{
  \institution{The Hong Kong University of
Science and Technology (Guangzhou)}
  \city{Guangzhou}
  \country{China}}
\email{hzhou269@connect.hkust-gz.edu.cn}

\author{Peiran Liu}
\affiliation{
  \institution{The Hong Kong University of
Science and Technology (Guangzhou)}
  \city{Guangzhou}
  \country{China}}
\email{pliu868@connect.hkust-gz.edu.cn}

\author{$\text{Hao Dong}^{\dag}$}
\affiliation{
  \institution{Peking University}
  \city{Beijing}
  \country{China}}
\email{hao.dong@pku.edu.cn}

\author{$\text{Yiding Ji}^{\dag}$}
\affiliation{
  \institution{The Hong Kong University of
Science and Technology (Guangzhou)}
  \city{Guangzhou}
  \country{China}}
\email{jiyiding@hkust-gz.edu.cn}

\begin{abstract}

Decision transformer based sequential policies have emerged as a powerful paradigm in offline reinforcement learning (RL), yet their efficacy remains constrained by the quality of static datasets and inherent architectural limitations. Specifically, these models often struggle to effectively integrate suboptimal experiences and fail to explicitly plan for an optimal policy. To bridge this gap, we propose \textbf{Imaginary Planning Distillation (IPD)}, a novel framework that seamlessly incorporates offline planning into data generation, supervised training, and online inference. Our framework first learns a world model equipped with uncertainty measures and a quasi-optimal value function from the offline data. These components are utilized to identify suboptimal trajectories and augment them with reliable, imagined optimal rollouts generated via Model Predictive Control (MPC). A Transformer-based sequential policy is then trained on this enriched dataset, complemented by a value-guided objective that promotes the distillation of the optimal policy. By replacing the conventional, manually-tuned return-to-go with the learned quasi-optimal value function, IPD improves both decision-making stability and performance during inference. Empirical evaluations on the D4RL benchmark demonstrate that IPD significantly outperforms several state-of-the-art value-based and transformer-based offline RL methods across diverse tasks.

\end{abstract}


\newcommand{\BibTeX}{\rm B\kern-.05em{\sc i\kern-.025em b}\kern-.08em\TeX}

\begin{document}


\pagestyle{fancy}
\fancyhead{}


\maketitle


\section{Introduction}
\begin{figure*}[htbp]\label{method_fig}  
    \centering
    \includegraphics[width=\textwidth]{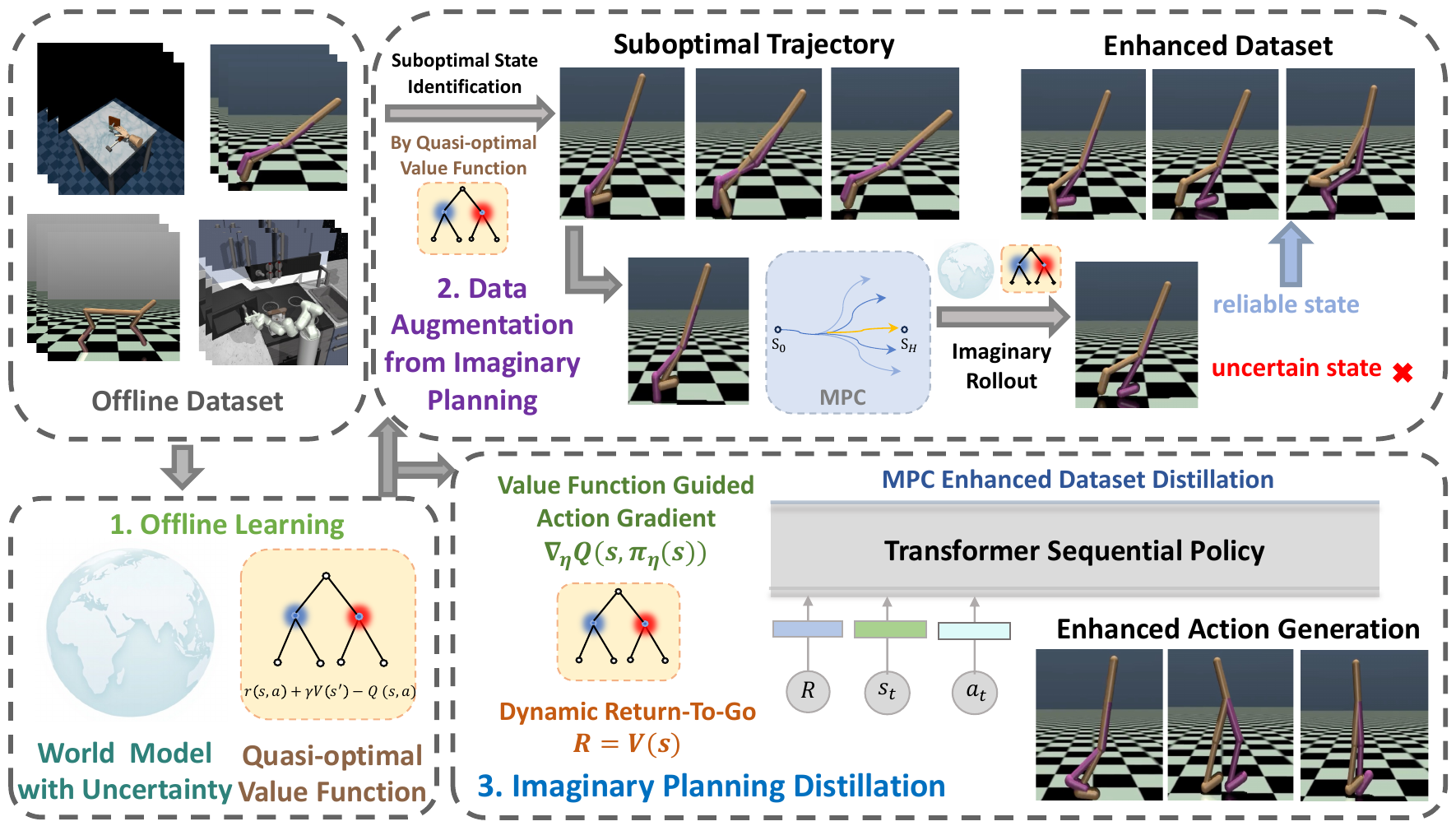}
    \vspace{-10pt}
    \caption{An overview of IPD. The process begins by learning a world model with uncertainty measure and a quasi-optimal value function from the original offline suboptimal dataset. Suboptimal states are identified using the value function, and their corresponding trajectories are replaced with imaginary rollouts generated via Model Predictive Control, using the learned world model and value function. Each generated trajectory is evaluated for uncertainty before incorporated into the enhanced dataset. Finally, a Transformer-based sequential policy is trained on this MPC-enhanced dataset, with additional supervision from the value function via action gradients and a dynamic return-to-go. By distilling the imaginary planning, which streamlines both MPC and dynamic programming, into the Transformer based policy, IPD enables the generation of superior actions.}
\end{figure*}


Reinforcement Learning (RL) has achieved remarkable success across various applications, ranging from robotic control~\cite{qin2025efficient, ibarz2021train,bai2025star,bai2025retrieval,Bai_Zhang_Tao_Wu_Wang_Xu_2023,Bai_Liu_Du_Wen_Yang_2025,wang2025adamanip,heng2025boosting}, autonomous driving~\cite{kiran2021deep} and chip design~\cite{mirhoseini2021graph} to complex strategic games~\cite{silver2016mastering, mnih2015human,wang2021tomc,ma2024fast}. However, the real-world deployment of online RL is often restrained by the high cost and safety risks associated with active exploration of under trained policies. As a safer alternative, offline RL enables policy training using fixed, pre-collected datasets without requiring environmental interactions~\cite{fujimoto2019off}. Despite this advantage, static datasets present significant challenges, most notably the value overestimation caused by state-action distribution shifts~\cite{kumar2020conservative}. To mitigate this issue, prior works have explored regularizing value function approximation or constraining policy deviation from the data collection policy.

Recently, a new class of offline RL algorithms, Decision Transformer~\cite{chen2021decision} and its variants, have emerged and gained prominence in language~\cite{achiam2023gpt,wolf2020Transformers} and vision~\cite{peebles2023scalable,dosovitskiy2020image} tasks. They leverage the Transformer's architecture, whose strong sequence modeling capabilities facilitate a reformulation of traditional Temporal Difference-based offline RL as a supervised conditional sequence generation problem. Although these models excel at modeling sequences, they rely on conditional sequence imitation and fundamentally lack the dynamic programming-based RL mechanisms, therefore struggling to stitch suboptimal trajectories into an optimal policy~\cite{brandfonbrener2022does}. Several approaches have been developed to address this limitation, including elastic context selection~\cite{wu2024elastic}, RL objective regularization~\cite{hu2024q,zhuang2024reinformer}, and return-to-go relabeling~\cite{yamagata2023q}. However, these methods offer only marginal performance improvements, as they do not fully integrate the core principles of planning into the whole training and inference cycles of Transformer-based sequential policies.

In this work, we propose \textbf{Imaginary Planning Distillation (IPD)}, a novel framework that integrates implicit dynamic programming and explicit model predictive control into both training and inference of Transformer-based sequential policy, boosting optimal trajectory generation. We term this approach ``imaginary planning" since both the MPC rollouts and value refinements are performed entirely within a learned world model, requiring no direct interaction with the real world.

As illustrated in Figure 1, our method consists of the following key steps. First, we train a world model with well defined uncertainty measure and a quasi-optimal value function using the original offline dataset. Next, we leverage the learned world model, value function and the reliable set derived from uncertainty measure, for the purpose of identify suboptimal trajectories. They are then replaced by imagined rollouts in a reliable set generated from model predictive control, which effectively enhances the dataset's quality. Subsequently, we train the Transformer policy with an additional value objective, encouraging actions that yield higher Q-values. Thus, the results of dynamic programming are implicitly distilled from the learned value function. During inference, the conditioning is based on our proposed quasi-optimal value function rather than conventional return-to-go, which automatically predicts the optimal return and contributes to more effective decision-making.

We evaluate IPD on the D4RL benchmark~\cite{fu2020d4rl} and the results demonstrate consistent performance improvements over state-of-the-art value-based and Transformer-based offline RL methods. Substantial ablation studies further validate the contributions of key components of IPD, including MPC-driven data augmentation, value-guided action imitation, and return-to-go prediction. Additionally, we analyze how volume of imaginary data augmentation affects policy performance and uncover a scaling law, providing insights into the effectiveness of our framework.

In summary, our main contributions are threefold:
1) We introduce Imaginary Planning Distillation (IPD), a novel framework that seamlessly integrates supervised sequence modeling with imaginary planning.
2) IPD incorporates both implicit dynamic programming and explicit model predictive control into training and inference of Transformer-based policy, enhancing optimal trajectory generation.
3) We conduct extensive experiments and ablation studies on the D4RL benchmark, validating the superior performance of IPD over existing offline RL methods.

\section{Related Work}
\textbf{Offline Reinforcement Learning.} 
Offline RL~\cite{levine2020offline, fujimoto2019off} breaks away from the traditional paradigm of online reinforcement learning and exploration, enabling policy learning solely from pre-collected datasets. However, classic off-policy value-based or actor-critic algorithms~\cite{van2016deep, mnih2013playing, haarnoja2018soft,schulman2017proximal} suffer from out-of-distribution (OOD) issues, where the value function tends to overestimate unseen state-action pairs~\cite{kumar2020conservative,levine2020offline, fujimoto2019off}. To address this challenge of overestimation, the mainstream offline RL algorithms fall primarily into two categories: policy constraint and value regularization.

For policy constraint methods, they enforce an additional regularization term that measures the policy discrepancy between the learned policy and the behavior policy using different distance metrics, such as batch constraints\cite{fujimoto2019off}, KL divergence~\cite{wu2019behavior}, MMD distance~\cite{kumar2019stabilizing}, and MSE constraints~\cite{fujimoto2021minimalist}.
For value regularization methods, they assign lower values to OOD state-action pairs for the value function, mitigating the overestimation problem~\cite{kumar2020conservative, kostrikov2021offline}.

\textbf{Transformer-based Sequential Policy.}
More recently, in contrast to the previous approaches, Decision Transformer (DT)~\cite{chen2021decision} introduces a supervised sequence modeling paradigm by directly maximizing action likelihood.
The Transformer’s sequence modeling capabilities enable a reformulation of traditional Temporal Difference-based offline RL as a supervised conditional sequence generation problem.
However, despite its strength in sequence modeling, the Transformer struggles to stitch suboptimal trajectories into an optimal policy\cite{brandfonbrener2022does}, as it primarily relies on conditional sequence imitation rather than dynamic programming-based RL. To address this limitation, several techniques have been proposed, including elastic context selection\cite{wu2024elastic}, RL objective regularization\cite{hu2024q,zhuang2024reinformer}, and return-to-go relabeling\cite{yamagata2023q}. However, these approaches yield only limited performance gains, as none fully incorporate planning principles throughout the training and inference process of Transformer-based sequential policies. In contrast, our proposed IPD manages to integrate a comprehensive planning distillation for the Transformer-based sequential policy.

\textbf{Model-based reinforcement learning}. Model-based RL usually improves sample efficiency and generalization capacity by planning with learned dynamics models~\cite{janner2019trust, hansen2023td, hansen2022temporal} or augmenting data using model-free methods~\cite{pitis2022mocoda, laskin2020reinforcement}. Recent efforts have extended these benefits to offline settings.

ROMI~\cite{wang2021offline} uses a reverse dynamics model to generate rollouts reaching target goals, while MOCODA~\cite{pitis2022mocoda} introduces counterfactual transitions through locally factored models to address out-of-distribution generalization. MOBILE~\cite{sun2023model} incorporates uncertainty quantification for conservative model usage, and SUMO~\cite{qiao2025sumo} estimates uncertainty via search-based cross-entropy alignment with in-distribution data. TD-MPC~\cite{hansen2023td, hansen2022temporal} applies latent dynamics modeling and trajectory optimization for online control.

In contrast, IPD integrates dynamic programming-based value learning with MPC planning to synthesize stitched imaginary trajectories from offline data. These trajectories are filtered using uncertainty-aware checks and used to enhance Transformer-based sequential policy learning. To our knowledge, IPD is the first framework to combine model-based MPC and dynamic programming for implicit trajectory stitching, enabling Transformer policies to exceed the limitations of the original offline dataset.

\section{Method}

This section proposes Imaginary Planning Distillation (IPD), an offline reinforcement learning framework to improve Transformer based sequential policy learning by distilling imaginary planning through uncertainty-aware data augmentation and value based guidance. The process is structured into four distinct phases: learning a quasi-optimal foundation, equipping the word model with an uncertainty measure, augmenting dataset via MPC, and distilling imaginary planning knowledge into the final Transformer policy.

We begin by introducing a quasi-optimal value function, which is learned from the original offline dataset using offline Q learning, and grounded in the principles of dynamic programming. In parallel, we train a reliable world model with uncertainty estimation, which is employed in the subsequent data augmentation stage. Next is the data augmentation phase. The learned value function is used to identify suboptimal trajectories, which are then replaced with high-quality trajectories generated using model predictive control (MPC). This process is guided by both the world model and the value function. Each generated trajectory is evaluated using the uncertainty estimate from the world model before being added to the enhanced dataset. 
Finally, the Transformer based sequential policy is distilled from imaginary planning, consisting of three core components: the MPC enhanced dataset, action gradients guided by the value function, and a dynamic return to go. The ``imagined" MPC rollouts are generated through the learned world model and the value function is shaped by offline Q learning, which implicitly incorporates the principle of dynamic programming. Distilling these components into the final policy mitigates the influence of suboptimal supervision and leads to improved policy performance.


\subsection{Offline Quasi-Optimal Value Function Learning}

IPD initiates by establishing a robust value function to mitigate the overestimation on out-of-distribution state-action pairs common in offline Q learning. We follow the principle of
Implicit Q Learning (IQL)\cite{kostrikov2021offline}, which restricts the Bellman update within the support of the dataset distribution. Specifically, we replace the original expectile regression with a Huber-expectile regression in Eq. \ref{huber_regess}, which provides asymmetric weighting for optimal value estimation and is more resilient to dataset outliers~\cite{hao2007quantile}. Here, $\delta$ is a hyperparameter that controls the transition between quadratic and linear loss.

Given the original offline dataset $D$, we derive a quasi-optimal value function $V_\psi(s)$ and Q function $Q_\theta(s, a)$ by optimizing the objectives in Eq.~(\ref{iql_value}) and Eq.~(\ref{iql_Q}), respectively. The target Q function is denoted as $Q_{\hat{\theta}}(s, a)$. The Huber loss can be represented as follows

\begin{equation}
\label{huber_regess}
\mathcal{L}_{\text{Huber}}^\delta(\textbf{e}_V) = 
\begin{cases}
\frac{1}{2} \textbf{e}_V^2, & \text{if } |\textbf{e}_V| \leq \delta, \\
\delta \left( |\textbf{e}_V| - \frac{1}{2} \delta \right), & \text{otherwise},
\end{cases}    
\end{equation}
which reduces the influence of large errors by transitioning from a quadratic to a linear penalty beyond a threshold $\delta$. Specifically, the Huber loss above is expressed using expectile regression as follows:

\begin{equation}
\label{huber_exp}
\mathcal{L}_2^{\tau, \text{Huber}}(\textbf{e}_V) = |\tau - \mathbb{I}(\textbf{e}_V < 0)| \cdot \mathcal{L}_{\text{Huber}}^\delta(\textbf{e}_V).
\end{equation}

This formulation preserves the asymmetric weighting of expectile regression while improving robustness to outliers, where the parameter $\tau$ controls the degree to which the agent is encouraged to learn the optimal action-value function
\begin{equation}
    \label{iql_value}
\mathcal{L}_V(\psi) = \mathbb{E}_{(s, a) \sim \mathcal{D}}\left[\mathcal{L}_{\text{Huber}}^\delta\left(Q_{\hat{\theta}}(s, a)-V_\psi(s)\right)\right] 
\end{equation}

\begin{equation}
    \label{iql_Q}
\mathcal{L}_Q(\theta) = \mathbb{E}_{(s, a, s') \sim \mathcal{D}}\left[\left(r(s,a)+\gamma V_\psi(s')-Q_\theta(s,a)\right)^2\right]
\end{equation}

Using $Q_\theta(s, a)$ and $V_\psi(s)$, we derive a quasi-optimal policy $\pi^{\textbf{QOP}}_\omega$ modeled as a Gaussian distribution, via advantage-weighted regression.The policy is optimized using Eq.~(\ref{iql_pi}), where $\beta \in [0, \infty)$ denotes the inverse temperature. As $\beta$ increases, $\pi^{\textbf{QOP}}_\omega$ becomes increasingly aligned with the action that maximizes the Q function.

\begin{equation}
\label{iql_pi}
\begin{aligned}
    \mathcal{L}^{\textbf{QOP}}_\pi(\omega) &= \mathbb{E}_{(s, a) \sim \mathcal{D}} \Bigg[ \exp \left( \beta \left( Q_{\hat{\theta}}(s, a) - V_\psi(s) \right) \right)
     * \log \pi^{\textbf{QOP}}_\omega(a \mid s) \Bigg]
\end{aligned}
\end{equation}
With $\beta \to \infty$, we extract a policy that maximizes the Q-values \cite{kostrikov2021offline}.
\begin{equation}
    \label{approx_max_q}
\pi^{\textbf{QOP}}_\omega( \arg\max_{a'} Q_{\theta}(s, a') \mid s) \approx 1.
\end{equation}

\subsection{World Model with Uncertainty Measure}
To facilitate effective data augmentation through model predictive contro (MPC), a world model is required to support the generation of imaginary rollouts. In our framework, it simultaneously learns a dynamic model $\mathcal{F}_{\phi}\left(\hat{{s}}_{t+1} \mid s_t, a_t\right)$ and a reward model $\mathcal{R}_{\phi}\left(\hat{{r}}_{t+1} \mid s_t, a_t\right)$.

In the offline setting, a standard dynamics model may fail to accurately capture the true system behavior due to limited coverage of the state-action space and inherent stochasticity in the transitions. Such errors will accumulate and degrade the quality of the generated trajectories during imaginary planning. Using a probabilistic ensemble (PE) world model to mitigate this issue, IPD explicitly models both aleatoric uncertainty arising from environmental randomness, and epistemic uncertainty stemming from limited knowledge or training data. Specifically, the next-state world model is implemented as an ensemble of Gaussian mixture models:
\begin{equation}
    \label{world_model}
\begin{split}
\tilde{\mathcal{F}}_{\mathbf{PE}}\left(\hat{{s}}_{t+1} \mid {s}_t, {a}_t\right) = \frac{1}{E} \sum_{e=1}^E\mathcal{F}_{{\phi}_{e}}\left(\hat{{s}}_{t+1} \mid {s}_t, {a}_t\right);\\
\mathcal{F}_{{\phi}_{e}}\left(\hat{{s}}_{t+1} \mid {s}_t, {a}_t\right) = \mathcal{N}\left({\mu}_{{\phi}_{e}}({s}_t, {a}_t), {\Sigma}_{{\phi}_{e}}({s}_t, {a}_t)\right)
\end{split}
\end{equation}
where $e \in {1, \ldots, E}$ indexes the ensemble members, and ${\Sigma}{{\phi}_{e}}(s_t, a_t)$ captures the aleatoric uncertainty. In addition to this, epistemic uncertainty is quantified by measuring the disagreement among ensemble members. Normally, this is computed as the Kullback-Leibler (KL) divergence between the ensemble's Gaussian mixture model $\tilde{\mathcal{F}}_{\mathbf{PE}}$ and each individual model distribution ${\mathcal{F}}_{\phi_e}$.

Unfortunately, the KL divergence between the Gaussian mixture models does not have a closed-form solution. To reinstate computational tractability during imaginary rollouts with MPC, we introduce a geometric Jensen-Shannon (GJS) divergence~\cite{frauenknecht2024trust, nielsen2019jensen} based uncertainty measure $\mathcal{U}$, which provides a tractable form for model disagreement. The measure decomposes the conventional KL divergence between the ensemble and individual models into the following iterative computations:


\begin{equation}
\begin{aligned}
    \mathcal{U}(s, a) &= \frac{1}{E(E-1)} \sum_{(i,j)\in\mathcal{P}} \mathcal{J}_{\mathrm{GJS}}(\mathcal{N}_i \| \mathcal{N}_j), \\
    &\text{where } \mathcal{P} \triangleq \{(i,j) \mid 1 \leq i < j \leq E\}
\end{aligned}
\end{equation}

\begin{equation}
\label{gjs}
\mathcal{J}_{\mathrm{GJS}}(\mathcal{N}_i | \mathcal{N}_j) = \frac{1}{2}\left[\mathcal{J}_{\mathrm{KL}}(\mathcal{N}_i | \mathcal{N}_{ij}) + \mathcal{J}_{\mathrm{KL}}(\mathcal{N}_j | \mathcal{N}_{ij})\right]
\end{equation}

\begin{equation}
\mathcal{N}_{ij} \sim (\mu_{ij}, \Sigma_{ij}), \quad 
\Sigma_{ij} = \left( \tfrac{1}{2}\Sigma_{\phi_i}^{-1} + \tfrac{1}{2}\Sigma_{\phi_j}^{-1} \right)^{-1}
\end{equation}

\begin{equation}
\mu_{i j}=\Sigma_{i j} \left(\frac{1}{2}\left(\Sigma_{\phi_{i}}\right)^{-1} \mu_{\phi_i}+\frac{1}{2}\left(\Sigma_{\theta_j}\right)^{-1} \mu_{\theta_j}\right)
\end{equation}

Based on the uncertainty measure $\mathcal{U}(s, a)$, we then define a threshold parameter $\kappa$ to filter a reliable subset from a generated rollout set $S$, as defined in Eq.~(\ref{uncertainty_set}). This procedure evaluates the reliability of imagined rollouts, ensuring that only trajectories with acceptable uncertainty levels are included in the augmented dataset.
\begin{equation}
    \label{uncertainty_set}
\mathcal{E} = \{ (s,a) \in \mathcal{S} \mid \mathcal{U}(s, a)  < \kappa\}
\end{equation}

Although ensemble models with Gaussian outputs are effective at capturing uncertainty, they also suffer from training instability due to variance collapse or explosion. To mitigate this issue, we employ Gaussian reparameterization to compute the predicted next state of the ensemble model, denoted by $\mu_\mathbf{PE}$. Additionally, we introduce an exponential decay schedule to regularize the predicted covariance $\Sigma_\mathbf{PE}$ during training. These strategies are incorporated into the training loss to stabilize learning. The resulting state-transition consistency loss $\mathcal{L}_{c}$ is formulated as follows:

\begin{equation}
    \label{gaussian_rep}
    \begin{aligned}
    \Sigma_{\mathbf{PE}} &=  \frac{1}{E}\sum_e \Sigma_{\phi_e}\\\mu_{\mathbf{PE}} &= \frac{1}{E}\sum_e\mu_{\phi_e} + \Sigma_{\mathbf{PE}}^{\frac{1}{2}} \cdot \epsilon, \epsilon \sim \mathcal{N}(0,I)
    \end{aligned}
\end{equation}

\begin{equation}
\label{world_model_loss}
\begin{aligned}
\mathcal{L}_{c} &= \frac{1}{|\mathcal{D}|}\sum_{\substack{(s_t,a_t,s_{t+1}) \in \mathcal{D}}} 
    \left\|\mu_{\mathbf{PE}}(s_t,a_t) - s_{t+1}\right\|^2 \\
&\quad + \left\|\gamma_{\text{exp}}(\Sigma_{\mathbf{PE}}(s_t,a_t) - \Sigma_{\text{reg}})\right\|^2
\end{aligned}
\end{equation}

The above covariance regularization is introduced mainly to stabilize training. The predicted covariance $\Sigma_{\mathbf{PE}}$ is guided towards a reasonable regularization target $\Sigma_{\text{reg}}$ during the initial training stage, under the control of a designed exponentially decayed regularization weight $\gamma_{\text{exp}}$. The decay factor $\gamma_{\text{exp}}(k)$ is defined as:

\begin{equation}
\label{exp_decay}
\gamma_{\text{exp}}(k) = \gamma_0 \cdot \exp\left(-\frac{k}{T}\right),
\end{equation}
where $k$ is the current training iteration step,
 $\gamma_0$ is the initial regularization scale, $T$ is the decay rate controlling how quickly the regularization diminishes steps. In addition, the loss of the reward model and the total loss of the world model can be calculated as
\begin{equation}
    \label{reward_loss}
\begin{split}
&\mathcal{L}_r = \frac{1}{|\mathcal{D}|}\sum_{\substack{(s_t,a_t,r_{t+1})  \in \mathcal{D}}}\left| \mathcal{R}_{\phi}(s_t, a_t) -r_{t+1} \right|^2, \mathcal{L}_{\text{world}} = \alpha_c \mathcal{L}_c +\alpha_r \mathcal{L}_r    
\end{split}
\end{equation}
For more details regarding the implementation for $\gamma_{\text{exp}}$, $\Sigma_{\text{reg}}$, $\alpha_c$ and $\alpha_r$, please refer to Experiments section.

\subsection{Data Augmentation with Imaginary Planning}
Building upon the quasi-optimal value function and the world model with uncertainty estimation, the next phase of IPD leverages these learned components to generate reliable data through MPC-based imaginary planning, which identifies and subsequently replaces suboptimal segments within the dataset.

The imaginary planning procedure consists of two key stages. The first stage, referred to as \textbf{Suboptimal State Identification}, aims to identify state-action pairs in the dataset that are likely to benefit most from enhancement. These are typically suboptimal samples where better policies yield improved returns. The identification is based on evaluating the discrepancy between the real return observed in the dataset and the return that could be obtained through imaginary rollouts.
Specifically, let $H_I$ denote the imaginary planning horizon. For each state in a trajectory, the imaginary return $R_{\text{Imagine}}$ and the real return $R_{\text{Real}}$ are computed as follows:
\begin{equation}
\label{img}
R_{\text{Imagine}}(s_t) = \sum_{k=0}^{H_I-1} \gamma^k \cdot 
\mathcal{R}_\phi(\hat{s}_{t+k}, \pi^{\textbf{QOP}}_\omega(\hat{s}_{t+k}))+V_\psi(\hat{s}_{H_I})
\end{equation}
\begin{equation}
\label{real}
R_{\text{Real}}(s_t) = \sum_{k=0}^{H_I-1} \gamma^k \cdot r(s_{t+k}, a_{t+k}) + V_\psi({s}_{H_I}) 
\end{equation}
where $R_{\text{Imagine}}$ is computed by rolling out the quasi-optimal policy $\pi^{\textbf{QOP}}_\omega$ for $H_I$ steps using the learned world model to simulate future states $\hat{s}$. In contrast, $R_{\text{Real}}$ is derived from the original trajectory stored in the offline dataset. By comparing these two returns, as defined in Eq.~(\ref{img}) and Eq.~(\ref{real}), IPD is able to identify candidate states within trajectories that significantly underperform relative to the imagined quasi-optimal rollouts. These states are selected for enhancement in the next phase.

The core principle of this stage is to select the top-$K$ states ranked by their potential improvement, measured as the difference $R_{\text{Imagine}} - R_{\text{Real}}$. The resulting set of selected states, denoted as $S_e$, represents those with the highest potential for enhancement and are thus prioritized for data augmentation.

Fundamentally, the main principle for the \textbf{Suboptimal State Identification} stage is to select states with the greatest value differences $R_{\text{Imagine}}(s_t) - R_{\text{Real}}(s_t)$.
To ensure sufficient context for the transformer policy while reserving space for imaginary rollout, we process trajectories with asymmetric windowing: each candidate state $s_t$ maintains $H_{\text{con}}$ historical states for conditioning while preserving $H_{I}$ future steps for augmentation, forming segments:

\begin{equation}
\label{window}
\tau_{\text{window}} = \{s_{t-H_{\text{con}}}, ..., s_t, ..., s_{t+H_{I}}\}.
\end{equation}
where the left $H_\text{con}$ states provide conditioning and the right $H_{I}$ states accommodate MPC-generated augmentation.

Given an original trajectory $\tau_o$ from dataset $D$, we segment it using a sliding window with step size 1, as defined in Eq.~(\ref{window}). The window ensures that the beginning of the trajectory contains $H_\text{con}$ historical states and the end contains $H_I$ future steps. The state at the center of each segment, i.e., the current state $s_t$, is added to the candidate state set $S_o$. We then compute the value difference for each candidate state and rank them in descending order as follows:

\begin{equation}
\mathcal{S}_{\text{sorted}} = \operatorname{argsort}_{s_t \in \mathcal{S}_{o} } \left[ R_{\text{Imagine}}(s_t) - R_{\text{Real}}(s_t) \right].
\end{equation}

We then select the top $K$ states for augmentation, where $K$ is determined by the augmentation ratio $N_\text{aug}$, as shown in Eq.~(\ref{equ:augratio}):

\begin{equation}\label{equ:augratio}
K = \min\left(\left\lfloor \frac{N_{\text{aug}} \cdot |\tau_o|}{H_{I}} \right\rfloor, |\mathcal{S}_{\text{sorted}}|\right),
\end{equation}
The final set of selected suboptimal states for augmentation is:

\begin{equation}
\mathcal{S}_{\text{aug}} =  \mathcal{S}_{\text{sorted}}[1:K]
\end{equation}

For each selected state $s_t^{(i)} \in \mathcal{S}_{\text{aug}}$, we generate an augmented trajectory segment of length $H_{I}$ with MPC and world model:

\begin{equation}
\tau_{\text{aug}}^{(i)} = \left\{s_t^{(i)}, \hat{s}_{t+1}^{(i)}, ...,\hat{s}_{t+H_I-1}^{(i)} \right\},
\end{equation}
where $\hat{s}_{t+k}$ refers to the generated imaginary rollout state. The rollout action $a_{t+k}^{(i)}$ is obtained through MPC planning:

\begin{equation}
a_{t+k}^{(i)} = \pi_{\text{mpc}}^*(s_{t+k}^{(i)}) \quad \text{for} \quad k = 0,...,H_{I}-2.
\end{equation}
All generated state-action pairs in the MPC planning stage must satisfy the uncertainty constraint.

Note that $H_{I}$ (data augmentation horizon) differs from the MPC planning horizon $H_m$. 
$H_{I}$ represents synthetic rollout length while $H_m$ determines optimal action selection depth.

After identifying the states that require enhancement, IPD proceeds to the next stage by performing reliable optimal planning via MPC for these selected candidate states. This step leverages both the previously learned world model with uncertainty estimation and the quasi-optimal value function and policy.

To ensure reliability, IPD constrains the MPC rollouts to remain within the uncertainty set $\mathcal{E}$ defined in Eq.~(\ref{uncertainty_set}). This constraint guarantees that optimal actions are selected only from model-confident regions of the state space, thereby reducing the risk of compounding model errors during imaginary data generation.

For optimal planning, IPD samples $N_{\text{mpc}}$ candidate trajectories of horizon $H_m$ using the learned world model. Among these, the action of the first step of the trajectory with the highest cumulative discounted return is selected:
\begin{equation}
\label{mpc_sample}
\begin{split}
\pi_{mpc}^*(a_t \mid s_t) =&\underset{\{a_t^{(i)}\}_{i=1}^{N_{\text{mpc}}}}{\arg\max} \sum_{k=0}^{H_m} \gamma^k \mathcal{R}_\phi\left(\hat{s}_{t+k}^{(i)}, \pi^{\textbf{QOP}}_\omega(\hat{s}_{t+k}^{(i)})\right) \\ &+\gamma^{H_m}V_\psi(\hat{s}^{(i)}_{t+H_m})
\end{split}
\end{equation}
where $\hat{s}_{t+k}^{(i)}$ denotes the $k$-th state in the $i$-th sampled trajectory, predicted by $\tilde{\mathcal{F}}_{\mathbf{PE}}$. The pseudocode for the imaginary data augmentation procedure is provided in Algorithm~\ref{alg:mpc_data_gen}.

\subsection{Imaginary Planning Distillation}

Based on the previously learned quasi-optimal value function and the augmented dataset, we further distill the knowledge obtained from these imaginary planning into the Transformer based sequential policy. This distillation process of IPD streamlines three key components. First, the learned quasi-optimal Q function provides action gradient guidance in the form of $\nabla_\eta Q(s, \pi_\eta(s))$, serving as a regularization signal to refine the policy. Second, the quasi-optimal value function dynamically guides the Transformer's return-to-go estimation, allowing it to infer the potential of future rewards directly from state inputs, without the need to manually define fixed target values. Third, the augmented dataset incorporates high-quality trajectories generated by MPC planning, which improves the overall performance of action supervision.

The joint effect of these components is captured by the total IPD loss function, which integrates both sequence modeling and quasi-optimal $Q$ function guided regularization:

\begin{equation}
\begin{split}
\mathcal{L}_{\mathrm{IPD}}(\eta) = & \mathbb{E}_{(s_t, a_t) \sim \mathcal{D}_{\text{aug}}} \Bigg[ \underbrace{\left( a_t - \pi_{\eta} \left( \{s, a\}_{t-K:t-1}, V_\psi(s_t), s_t \right) \right)^2}_{\text{Sequence Modeling Term}} \\
& - \alpha \cdot \underbrace{Q_{\theta} \left( s_t, \pi_{\eta} \left( \{s, a\}_{t-K:t-1}, V_\psi(s_t), s_t \right) \right)}_{\text{Q-Value Regularization Term}} \Bigg].
\end{split}
\end{equation}

\noindent where $\mathcal{D}_{\text{aug}}$ is the augmented dataset constructed in Alg.~\ref{alg:mpc_data_gen} and contains trajectories $\tau_i$.
$(s_t, a_t)$ is a state-action pair sampled from $\mathcal{D}_{\text{aug}}$.
$\{s, a\}_{t-K:t-1}$ represents the context window of past $K$ states and actions.
$V_\psi(s_t)$ is the estimated state value from the quasi-optimal value function, providing a dynamic prompt for return-to-go estimation.
$\pi_{\eta}(\cdot)$ denotes the Transformer policy parameterized by $\eta$.
$\alpha$ is a weighting coefficient that balances the two loss components.

This composite loss function ensures that the policy not only replicates the high-quality actions from the augmented dataset but is also regularized by the Q-function. Thus, actions are produced and \textit{outperform} those in the dataset, facilitating performance improvement and knowledge distillation from the imaginary planner.


\begin{algorithm}[ht!]
\caption{\textbf{Imaginary Data Augmentation}}
\label{alg:mpc_data_gen}
\begin{algorithmic}[1]
\REQUIRE Pretrained model: $V_\psi$, $\pi^{\textbf{QOP}}_{\omega}$, $\tilde{\mathcal{F}}_{{\mathbf{PE}}}$ , $\mathcal{R}_{\phi}$
\REQUIRE States set: $\mathcal{E}$, $S_e$ 
\REQUIRE horizon $H_I$, MPC horizon $H_m$, MPC rollout number $N_{mpc}$
\vspace{-3mm}
\STATE \textbf{Initialize:} $\mathcal{D}_{\text{aug}} \leftarrow \emptyset$

\FOR{each state $s_0 \in S_e$}
    \STATE $s \leftarrow s_0$
    \STATE $\tau \leftarrow [\,]$
    
    \FOR{$t \leftarrow 1$ \TO $H_I$}
        \IF{$s \not\in \mathcal{E}$} 
        \STATE \textbf{break}        
        \quad \quad // Uncertainty check 

    \ENDIF
        
        \STATE \textbf{MPC Planning:}
        \STATE $a^* \leftarrow \text{MPC}(s, H_m,N_{mpc},\tilde{\mathcal{F}}_{{\mathbf{PE}}},\mathcal{R}_{\phi}, V_\psi)$  Eq. (\ref{mpc_sample}) 
        \STATE $\hat{r}, \hat{s}' \leftarrow \mathcal{R}_\phi(s, a^*), \tilde{\mathcal{F}}_{{\mathbf{PE}}}(s, a^*)$
        
        \STATE $\tau \leftarrow \tau \cup \{(s, a^*, \hat{r}, \hat{s}')\}$
        \STATE $s \leftarrow \hat{s}'$
    \ENDFOR
    
    \IF{$\tau \neq [\,]$}
        \STATE $\mathcal{D}_{\text{aug}} \leftarrow \mathcal{D}_{\text{aug}} \cup \{\tau\}$
    \ENDIF
\ENDFOR

\RETURN $\mathcal{D}_{\text{aug}}$
\end{algorithmic}
\end{algorithm}

\section{Experiments}

In this section, we present comprehensive experiments in D4RL benchmark \cite{fu2020d4rl}, which are designed to evaluate the performance of the proposed IPD in comparison with several baseline approaches. Specifically, we benchmark IPD against traditional Q-learning based methods, as well as recent Transformer based sequential policy methods, along with their improved versions that aim to address the problem of suboptimal trajectories. 

Complementary to those experiments, we conduct a series of ablation studies to assess the efficacy of different components of IPD. In particular, we explore how the superior planning quality of MPC compared to conventional greedy Q-learning strategies contributes to enhanced data generation, which highlight the advantages of MPC in handling complex decision spaces. Furthermore, we investigate how the scale of enhanced data generation influences the agent's ability to address complex tasks, with a focus on exploring potential scaling law in offline RL.
Additionally, we examine the influence of the quasi-optimal value function as a guiding mechanism, comparing it with manually engineered return-to-go functions. 

Our experiments are structured to investigate several critical research questions, each of which is explored in dedicated sections:

\begin{itemize}[leftmargin=*]
    \item Does IPD outperform Q-learning-based methods and conventional Transformer based sequential policy methods? 
    \item Does IPD surpass state-of-the-art advancements that overcome the suboptimal trajectory limitations of Transformer based sequential policy methods?
    \item Does MPC enhance the quality of data generation compared to vanilla greedy Q-learning methods? 
    \item What is the relation between the volume of data generation and the final performance of IPD?
    \item Does the quasi-optimal value function, as a guidance mechanism, mitigate the instability caused by manually engineered return-to-go and improve robustness? 
\end{itemize}

\begin{table*}
  \centering
  \scriptsize
  \setlength{\tabcolsep}{4pt}
  \small 
  \begin{tabular}{l|llllllllll}
    \hline
    \multicolumn{1}{c|}{\textbf{Gym Task}} & \multicolumn{1}{l}{IQL} & CQL & DT & \multicolumn{1}{r}{QDT} & DD & EDT & QT & Reinformer & \small\textbf{IPD} \\ \hline
    walker-medium & 78.3 & 83.0 & 74.0 & 67.1 & 82.5 & 72.8 & 87.6 & 80.5 & \textbf{89.5 $\pm$ 3.7} \\[1mm]

    walker-medium-replay & 73.9 & 77.2 & 79.4 & 58.2 & 68.9 & 74.8 & 94.2 &  72.9&\textbf{96.2 $\pm$ 2.1} \\[1mm] 
    hopper-medium & 66.3 & 69.4 & 67.6 & 66.5 & 79.3 & 64.5 & 78.0 & 63.5 & \textbf{81.6 $\pm$ 4.8} \\[1mm]
    hopper-medium-replay & 94.7 & 95.0 & 82.7 & 52.1 & 100.0 & 89.0 & 102.1 & 83.3 & \textbf{103.2 $\pm$ 3.7} \\[1mm]
    halfcheetah-medium & 47.4 & 49.2 & 42.6 & 39.3 & 49.1 & 42.5 & 49.1 & 42.9 & \textbf{51.2$\pm$2.6}  \\[1mm]
    halfcheetah-medium-replay & 44.2 & 45.5 & 36.6 & 35.6 & 39.3 & 37.8 & 48.9 & 39.0 & \textbf{49.9 $\pm$ 1.6}\\ \hline
    \multicolumn{1}{c}{\textbf{Kitchen Tasks}} \\ \hline
    kitchen-complete & 62.5 & 43.8 & 50.8 & 56.9 & 65.0 & 40.8 & 75.0 & 50.9 & \textbf{78.4 $\pm$ 4.6} \\[1mm]
    kitchen-partial & 46.3 & 49.8 & 57.9 & 58.3 & 57.0 & 10.0 & 73.2 & 73.1 & \textbf{74.3 $\pm$ 2.8} \\
 \hline
    \multicolumn{1}{c}{\textbf{Adroit Tasks}} \\ \hline
    hammer-human-v1 & 1.4 & 4.4 & 0.2 & 6.8 & 1.9 & 14.2 & \textbf{24.8} & 17.2 &  22.9 $\pm$ 2.3 \\
    pen-cloned-v1 & 37.3 & 39.2 & 75.8 & 64.2 & 42.8 & 48.4 & 90.1 & 82.4 & \textbf{92.8 $\pm$ 3.7}\\
 \hline
  \end{tabular}
\caption{Evaluating IPD and State-of-the-Art Q-Learning and Transformer-Based Sequential Policy Methods on D4RL Benchmarks: Performance Analysis Through Tenfold Episode Evaluations}
\label{main_result}
\end{table*}

\subsection{Baseline Methods}
We conduct a comprehensive comparison of IPD with several baseline methods, including traditional Q-learning-based approaches, such as Conservative Q-Learning (CQL) \cite{kumar2020conservative} and Implicit Q-Learning (IQL) \cite{kostrikov2021offline}, as well as vanilla Transformer-based sequential policy methods, notably the Decision Transformer (DT) \cite{chen2021decision}. Furthermore, we evaluate IPD against enhanced variants of DT. These methods either target Diffusion-Based Sequence Modeling or are specifically designed to overcome the challenge of suboptimal trajectory stitching, such as Decision Diffuser (DD) \cite{ajay2022conditional}, Elastic Decision Transformer (EDT) \cite{wu2024elastic}, Q learning decision Transformer (QDT) \cite{yamagata2023q}, Q-value regularized Transformer (QT) \cite{hu2024q} and Reinformer \cite{zhuang2024reinformer}. 


\subsection{Main Results}
This subsection summarizes the main results of our experiments carried out in three distinct domains: Gym tasks, Kitchen tasks, and Adroit tasks. These experiments cover ten tasks, as shown in Table \ref{main_result}, we performed $10$ evaluation episodes for each task.
To ensure a fair comparison and meaningful interpretation of the results, all scores have been normalized with respect to the D4RL benchmarks.

As illustrated in Table \ref{main_result}, IPD demonstrates superior performance compared to most offline Q-learning-based methods and  transformer-based approaches. This highlights IPD's ability to leverage the sequence modeling strengths of transformers while simultaneously enhancing its performance through the integration of dynamic programming techniques.

For Gym Tasks, in the context of these medium datasets, IPD excels by effectively harnessing MPC infused with value function to generate more high-quality data. 
Meanwhile, for Kitchen tasks that require effectively generalizing to novel states and long-horizon optimization and Adroit tasks that struggle in sparse human demonstration, IPD also demonstrates its advancements efficacy.
By incorporating data generation and quasi-optimal value function guidance, IPD leverages the implicit dynamic programming to guide the agent in producing higher-value actions, which results in a notable boost in effectiveness, enabling IPD to deliver outstanding performance and setting it apart from other methods that may struggle in scenarios with limited high-quality trajectories.

\subsection{Ablation Study}
This subsection presents a comprehensive ablation study on key components of IPD, emphasizing its data generation and quasi-value guidance. We analyze their contributions and impacts. 

\begin{figure}[ht!]
    \centering
    \includegraphics[width=8.5cm, height=6.5cm]{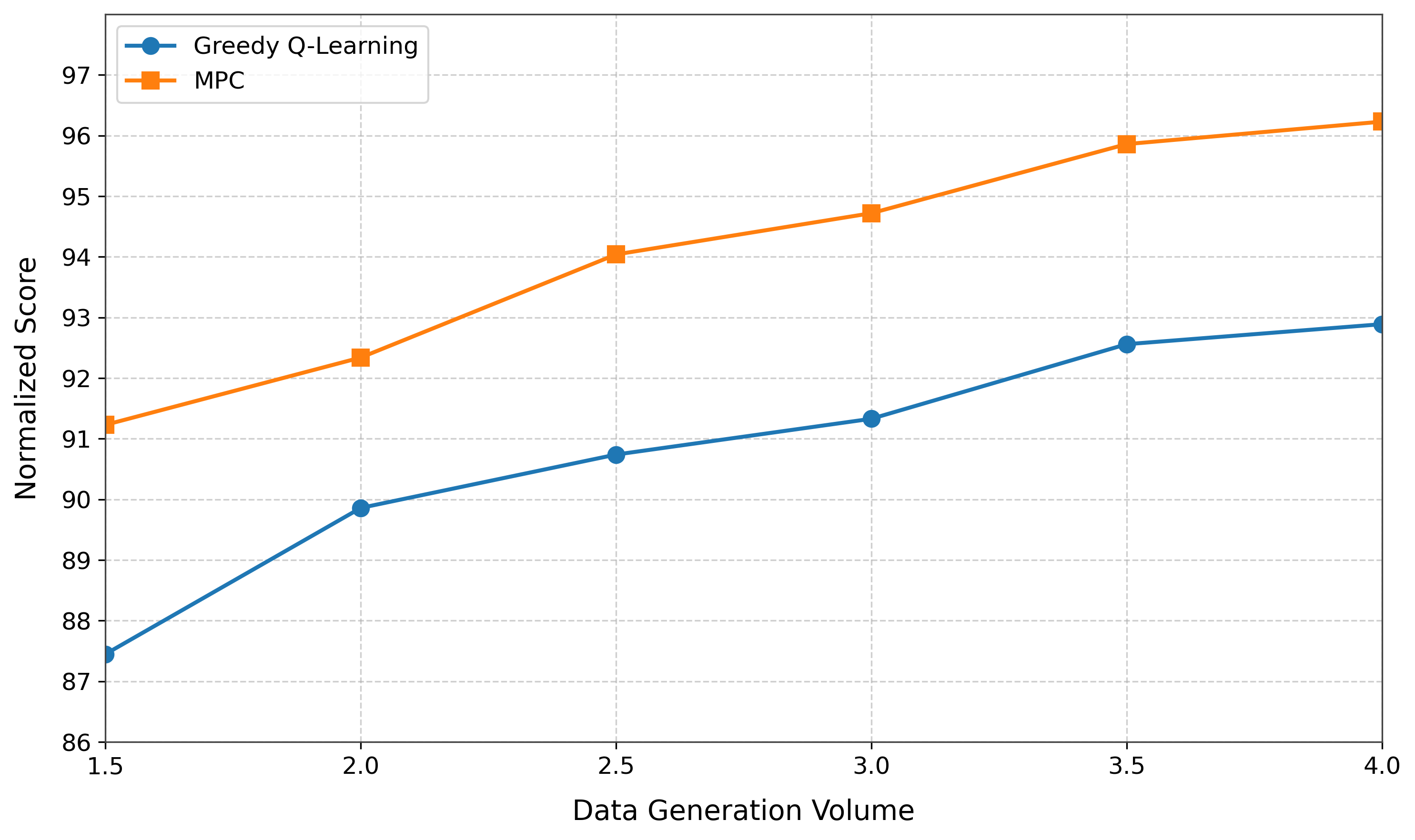}
    \vspace{-25pt}
    \caption{Comparison between MPC and Greedy Q-Learning data augmentation in Walker2d-medium-replay task.}
    \label{fig:mpc_vs_greedy}
\end{figure}

\textbf{Analysis for Data Generation.} We compare our MPC-based data generation procedure with a greedy Q-learning procedure, which utilizes the quasi-optimal policy in Eq. \ref{iql_pi} to directly generate action for rollouts.
Fig.~\ref{fig:mpc_vs_greedy} shows that data generation empowered by MPC surpasses vanilla greedy Q-learning in performance. 
By utilizing pretrained world model and quasi-value function, MPC can sample multiple trajectories and select the optimal one for action selection. By choosing the most optimal actions based on these sampled trajectories, MPC achieves a higher performance level. This approach not only demonstrates the effectiveness of our trained world model but also highlights MPC's ability to make more informed and strategic decisions.

In addition, the quantity of data generation reveals that as more data is generated, the performance in IPD shows an approximately linear improvement. This trend further demonstrates the effectiveness of our approach, and the data scaling effect in offline data generation. By utilizing world model and MPC for planning, we can generate better trajectories for the transformer-based sequential policy. This approach enhances the quality of the training data, leading to improved policy performance.

\vspace{-10pt}
\begin{figure}[ht!]
    \centering
    \includegraphics[width=8.5cm, height=6cm]{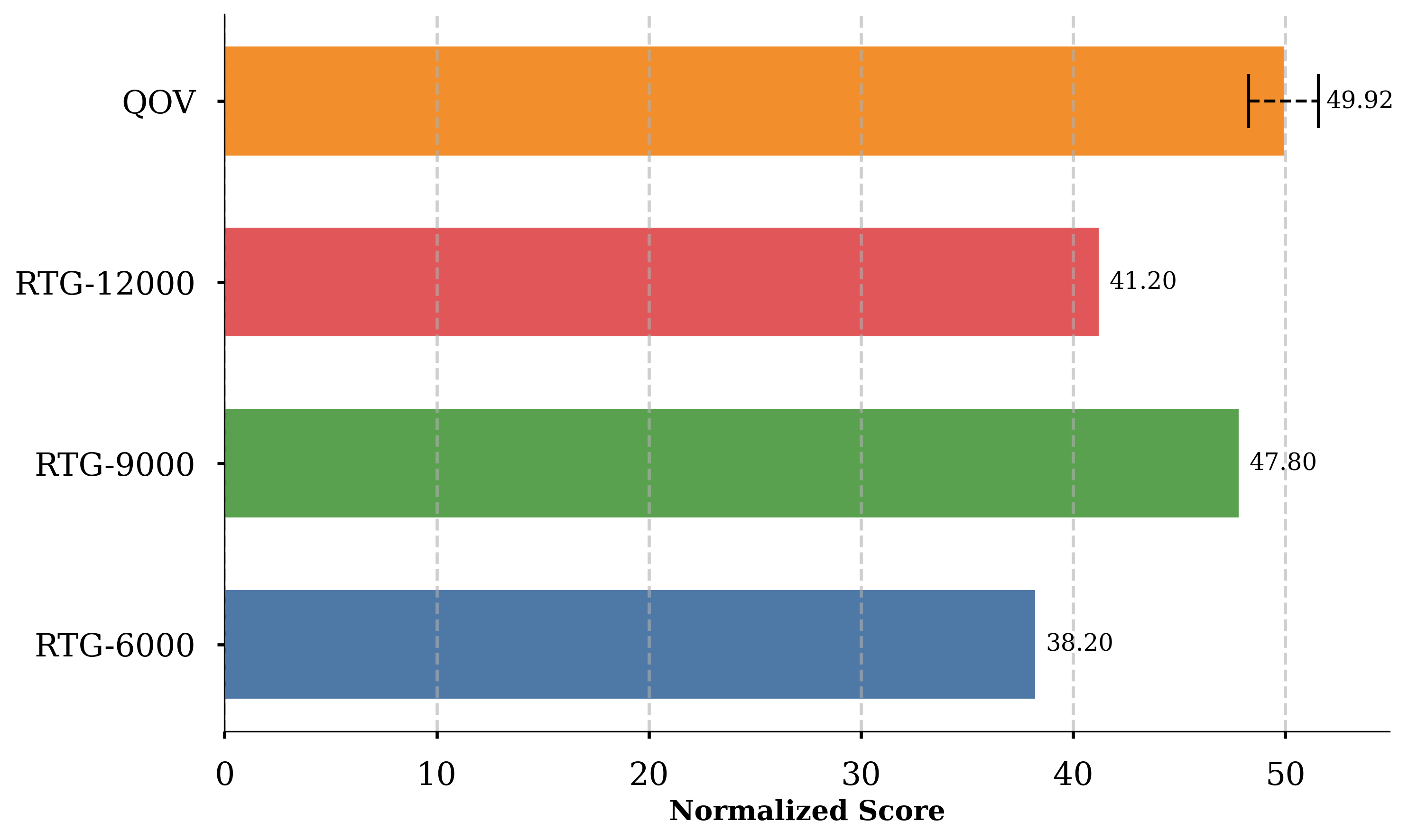}
    \vspace{-25pt}
    \caption{Performance Comparison between quasi-optimal value function and different setting of return-to-go in halfcheetah-medium-replay task.}
    \label{fig:quasivalue}
\end{figure}

\textbf{Analysis for Quasi-Value Function Guidance.} A critical challenge in Decision Transformers is the sensitivity to manually engineered Return-To-Go (RTG) values, which are computationally expensive during inference and result in instability of the algorithm's performance. This is primarily due to the arbitrary RTG values, which may not accurately reflect the optimal trajectory and cause suboptimal decision-making. We replace arbitrary RTG values with the learned Quasi-Optimal Value (QOV), which streamlines the inference process, eliminating the need for costly manual tuning while enhancing the robustness and stability of the algorithm.



As evidenced in Table \ref{main_result} and Fig. \ref{fig:quasivalue}, our approach yields a significantly lower variance in tests, indicating consistent and stable outcomes across different trials. In contrast, varying RTG values results in noticeable performance degradation and instability during inference, as the agent struggles to adapt to inconsistent guidance.

\subsection{Implementation Details}
In this subsection, we detail the implementation, including neural network architecture and hyperparameter settings for IPD, all experiments were conducted on NVIDIA GeForce RTX 4090 GPU.

\subsubsection{Quasi-optimal Value Function Learning }

We use the Adam optimizer with an initial learning rate of $3 \times 10^{-4}$. The Q, V, and actor networks each consist of two fully connected layers with 256 hidden units and ReLU activation functions. For the actor network, we apply a cosine annealing schedule to adjust the learning rate during training. To enhance training stability for the critic, we adopt a double Q-learning strategy. The policy is modeled as a Gaussian distribution with a state-independent standard deviation. For updating the target Q network, we use a soft update mechanism with a smoothing coefficient $a = 0.005$. The threshold for the Huber loss is set to the $\delta$-th percentile of the historically collected value error $\mathbf{e}_V$. Detailed hyperparameter settings are provided in Table~\ref{table_iql}.

\vspace{-5pt}
\begin{table}[ht!]
\centering
\begin{tabular}{cccc}  
\toprule
\textbf{Value Function} & \textbf{Value} & \textbf{World Model} & \textbf{Value}\\ 
\midrule
$\tau$             & 0.7     &   $\Sigma_{\text{reg}}$             & $e^{-4.5} \cdot I$      \\
$\beta$            & 3       &   $\gamma_0$             & 1.0      \\
$\delta$           & 0.96    &    $T$             & $10^5$     \\
$\gamma$           & 0.99    &    $\alpha_c$             & 0.5    \\
$\text{Dropout}$   & 0.01    &    $\alpha_r$             & 0.5      \\
\bottomrule
\end{tabular}    
\caption{Hyperparameters of Quasi-Optimal Value Function and World Model with Uncertainty Measure Learning.}
\label{table_iql}
\end{table}
\vspace{-5pt}

\subsubsection{World Model with Uncertainty Measure}
We use an ensemble of three models for world model, each of it is trained using the Adam optimizer with a learning rate of $3 \times 10^{-4}$. The architecture consists of multiple fully connected layers with 400 units, Layer Normalization, and ReLU activation functions to ensure stable gradient propagation. The dynamics and reward models share the same architecture but are trained independently. For the total loss of world model is $\mathcal{L}_\text{world} = \alpha_c \mathcal{L}_c +\alpha_r \mathcal{L}_r$, we set $\alpha_c = \alpha_r =0.5$ to maintain equal contribution of dynamic prediction and reward prediction. For exponential decay schedule, we constrain the covariance within $[e^{-10}, e^{0.2}]$.
Details are provided in Table~\ref{table_iql}.

\subsubsection{Data Generation with Imaginary Planning}
The specific parameter setting is listed in Table \ref{mpc}.

\begin{table}[ht!]
\centering
\begin{tabular}{cc}
\toprule
\textbf{Parameters} & \textbf{Value} \\ 
\midrule
$\gamma_\text{mpc}$           & 0.99          \\
$H_I$             & 10            \\
$H_m$            & 10              \\
$H_\text{con}$            & 10              \\
$N_\text{mpc}$           & 3           \\
\bottomrule
\end{tabular}
\caption{Hyperparameters of Imaginary Data Augmentation.}  
\label{mpc}
\end{table}

\begin{table}[ht!]
\centering
\begin{tabular}{@{}cc@{}}
\toprule
\textbf{Parameters} & \textbf{Value} \\ 
\midrule
$\text{Layers}$             & 4            \\
$\text{Dropout}$             & 0.01            \\
$\text{Embedding}$           & 256        \\
$\text{Attention heads}$           & 4    \\
$\text{Context length}$           & 20        \\
\bottomrule
\end{tabular}    
\caption{Hyperparameters of Sequential Policy}  
\label{trans}
\end{table}

\subsubsection{Transformer-based Sequential Policy}
For training the Transformer based sequential policy, we use the Adam optimizer with a learning rate of $3 \times 10^{-4}$ and employ the ReLU activation function. Detailed parameter settings are provided in Table~\ref{trans}.

\section{Conclusion}
This work developed Imaginary Planning Distillation (IPD), a novel framework that bridges the gap between supervised learning and reinforcement learning (RL). By integrating implicit dynamic programming with explicit model predictive control (MPC), IPD enables Transformer-based policies to transcend the limitations of suboptimal offline datasets. IPD utilizes an uncertainty-aware world model and a quasi-optimal value function to augment datasets with reliable imagined rollouts, significantly augmenting the dataset. In addition, we propose an optimal value gradient-guided action imitation objective that further integrates planning into the policy learning process. For inference, we replace the conventional return-to-go with a learned value function, facilitating dynamic guidance and stable decision-making. Our evaluation of IPD on the D4RL benchmark reveals consistent performance improvements over existing value-based and Transformer-based offline RL methods. Comprehensive ablation studies affirm the contributions of MPC-driven data augmentation, value-guided imitation, and learned return-to-go prediction. Additionally, our analysis of scaling laws offers valuable insights into the benefits of imaginary data augmentation.

In summary, IPD offers a comprehensive and principled approach that integrates ``imaginary" planning and sequence modeling for offline RL. Thus, this work paves the way for more effective policy learning in real-world decision-making tasks. 





\begin{acks}
\texttt{This work is partially supported by National Natural Science Foundation of China grants 62303389, 62373289; Guangdong Basic and Applied Basic Research Funding grants 2024A1515012586; Guangdong Scientific Research Platform and Project Scheme grant 2024KTSCX039 and Youth Talent Support Program of Guangdong Association for Science and Technology grant SKXRC2025463.}

\end{acks}



\bibliographystyle{ACM-Reference-Format} 
\bibliography{main}


\end{document}